\documentclass[11pt]{iopart}
\ioptwocol

\usepackage{iopams}

\usepackage{graphicx}
\usepackage{bm}

\usepackage[switch]{lineno} 

\begin{document}


\title[Suppressing simulation bias using multi-modal data]{Suppressing simulation bias in multi-modal data using transfer learning}

\author{Bogdan Kustowski$^1$, Jim A Gaffney$^1$, Brian K Spears$^1$, Gemma J Anderson$^1$, Rushil Anirudh$^1$, Peer-Timo Bremer$^1$, Jayaraman J Thiagarajan$^1$, Michael K G Kruse$^1$, Ryan C Nora$^1$}

\address{$^1$ Lawrence Livermore National Laboratory, 7000 East Ave, Livermore, CA 94550, USA}

\ead{kustowski1@llnl.gov}

\vspace{10pt}
\begin{indented}
\item[]March 2022
\end{indented}


\begin{abstract}
Many problems in science and engineering require making predictions based on few observations. To build a robust predictive model, these sparse data may need to be augmented with simulated data, especially when the design space is multi-dimensional. Simulations, however, often suffer from an inherent bias. Estimation of this bias may be poorly constrained not only because of data sparsity, but also because traditional predictive models fit only one type of observed outputs, such as scalars or images, instead of all available output data modalities, which might have been acquired and simulated at great cost. To break this limitation and open up the path for multi-modal calibration, we propose to combine a novel, transfer learning technique for suppressing the bias with recent developments in deep learning, which allow building predictive models with multi-modal outputs. First, we train an initial neural network model on simulated data to learn important correlations between different output modalities and between simulation inputs and outputs. Then, the model is partially retrained, or transfer learned, to fit the experiments; a method that has never been implemented in this type of architecture. Using fewer than 10 inertial confinement fusion experiments for training, transfer learning systematically improves the simulation predictions while a simple output calibration, which we design as a baseline, makes the predictions worse. We also offer extensive cross-validation with real and carefully designed synthetic data. The method described in this paper can be applied to a wide range of problems that require transferring knowledge from simulations to the domain of experiments. 
\end{abstract}


\let\thefootnote\relax\footnotetext{This work has been submitted to the IOPScience for possible publication. Copyright may be transferred without notice, after which this version may no longer be accessible.}

%
%
%
%
%


\section{Introduction}
\label{sec:intro}

Building predictive models is a common task in a wide range of disciplines ranging from science, engineering, and medicine, through finance, management, and insurance. If sufficient data quantity is available, an accurate, data-driven predictive model can be estimated. In many applications, however, only a handful of data points is available to determine model parameters and such models usually do not generalize well. If these sparse data samples additionally come from a multi-dimensional design space, a purely data-driven approach becomes unrealistic.

Alternatively, data can be modeled in a deterministic way using insights about the data generating process. Designing a perfect model is typically not possible because of insufficient understanding of the data, approximations necessary to make the problem tractable, or limited computational resources. Consequently, even the best available simulation cannot accurately predict the data.

With these challenges in mind, it is natural to consider a hybrid approach, where the information from sparse data is combined with the knowledge built into the simulation. To enable the hybrid approach, expensive simulations are often approximated by fast-to-evaluate, differentiable, and mutable surrogates~\cite{booker}~\cite{santnerbook}, which can predict simulation outputs from a predefined range of simulation inputs. Predictions $y_{surr}=S(x)$ of the simulation-trained surrogate model $S$ from the known inputs $x$, differ from the experimental outputs $y_{obs}$ by a bias term $\delta$.  Once the bias $\hat{\delta}$ has been estimated, the corrected predictions can be computed as:

\begin{equation} \label{eq:bayesian}
y_{corr} = S(x) + \hat{\delta}(x).
\end{equation}

A classic approach for finding the bias is Bayesian calibration~\cite{kennedyohagan}, where the bias is estimated simultaneously with any uncertain model parameters, the latter being the main objective of the calibration. This method is, however, not designed to work with a small number of observations and with diverse types of outputs, such as time series, images, or videos. With the current trend for deploying more sensors and more types of sensors in many aspects of our life~\cite{iot} and science~\cite{peternature}~\cite{walt}~\cite{microseismic}, as well as the advances in simulating such data~\cite{iotsim}~\cite{nora}~\cite{astro}~\cite{dassim}, there is a need to develop calibration methods that leverage all available data types. This is especially important when the observational data set is very sparse and more data types could break the non-uniqueness in the estimation.

Recently, an alternative approach for handling the simulation bias, called transfer learning, started gaining popularity. In principle, transfer learning is a general idea of leveraging the knowledge acquired in a source domain to improve a model that predicts data in a target domain. The method has been applied to various types of models including neural networks~\cite{caruna},~\cite{yosinski},~\cite{domainadaptation}, Markov logic networks~\cite{tlmc} and Gaussian processes~\cite{tlgp}. Researchers have started using this method to transfer the knowledge from the domain of simulations to the domain of real data~\cite{kustowski2019}~\cite{kellitl}~\cite{tlmanufacturing}, such that Equation \ref{eq:bayesian} is replaced with

\begin{equation} \label{eq:tl}
y_{corr} = S^{TL}(x, y_{obs}, S),
\end{equation}

where $S^{TL}$ is a new, transfer learned, predictive model. In this formulation, the bias $\delta$ is not explicitly estimated, in contrast to Equation \ref{eq:bayesian}. Instead, a bias-corrected model $S^{TL}$ is computed from the inputs $x$ and observed outputs $y_{obs}$, as a modification of the simulation-trained model $S$.  While~\cite{kustowski2019} and \cite{kellitl} have previously demonstrated that transfer learning calibration works for simple problems, where the observations consist of only scalar outputs, the real strength of this method, which has not been exploited so far, lies in the possibility of calibrating multi-modal data, thanks to recent advances in deep learning~\cite{rushilpnas}. One of the main contributions of this paper is including, for the first time, multi-modal output data to calibrate a simulation-trained surrogate using transfer learning. Incorporating multi-modal outputs cannot be achieved by simply replacing fully connected neural network layers with the convolutional ones; it requires a completely different and more complex model architecture, in which we have tested various transfer learning strategies for the first time.

To clarify the nomenclature, whenever we use the term ``calibration" in this paper, we refer to designing a corrected surrogate model $S^{TL}$ to better predict the observed outputs. This is different from the Bayesian calibration, whose main goal is to infer unknown parameters of a computer model, from which the unchanged model $S$ predicts the observed outputs, while allowing for some bias correction. In Section \ref{sec:simple}, we will also discuss the ``output calibration'' - a mapping between simulation and experimental outputs, which neither modifies the inputs nor the predictive model of a simulation.

For the machine learning community, it is worth clarifying that the problem we are trying to solve is different than the domain adaptation in image classification problems. We apply transfer learning to a regression problem, where image data are the outputs, not the inputs of the model. It is possible to view it as an example of few-shot learning~\cite{fewshot}, where only few observations are available for training. In few-shot learning, it is common to apply simple data augmentation techniques to add certain invariances in the model, typically for only one data modality. We take a much more comprehensive approach to build an augmented dataset by simulating all data modalities in a physically consistent way. During few-shot training, it is also common to apply model regularization. Transfer learning can be thought of as a sophisticated regularization technique. We will demonstrate that, unlike a standard $L2$ regularization, transfer learning constrains the solution space yielding a model that generalizes well despite being trained on only a handful of observations. It should be also noted that, in contrast to purely machine learning publications, where cutting-edge methods are typically tested on freely available benchmark datasets, our task involves predicting very expensive, hence limited in quantity, experimental and simulated data, which limits our cross-validation options. From the machine learning perspective, the novelty of this paper is the application of transfer learning to a new type of the regression problem without the existing benchmark, using a non-standard architecture.

Regarding the specific application of transfer learning discussed in this paper, all real data have been acquired during inertial confinement fusion (ICF) ~\cite{Atzeni}~\cite{bettihurricane} experiments at the National Ignition Facility (NIF) - the largest laser system in the world~\cite{NIFMiller}. During these experiments, which we also refer to as shots, the laser energy after being converted into X-rays, compresses a millimeter-scale target capsule. The capsule contains fuel consisting of two hydrogen isotopes: deuterium and tritium. Under extreme conditions caused by the compression, the nuclei of deuterium and tritium are fused into the nucleus of helium and a vast amount of energy is released. The goal of the experiments is to prove that ignition, where the released energy is larger than the energy provided by the laser, is possible in the laboratory setting. The successful demonstration would open up the path for building new types of nuclear reactors operating without heavy radioactive elements.  Developing more accurate predictive models is crucial for predicting new ICF experiments before they are carried out, for optimizing designs of new experiments~\cite{ovoid}~\cite{peterdesign}, and for better understanding of the physics controlling the performance of each experiment.

The paper is organized as follows. In Section \ref{sec:data}, we describe the ICF experimental and simulated  data, including the inputs and multi-modal outputs. In Section \ref{sec:method}, we first present the neural network architecture used to build the initial surrogate model of the simulations. Then, we discuss three different strategies of transfer learning in this architecture and select the preferred method. Presenting details of how these different strategies perform is beyond the scope of this paper. In Section \ref{sec:results}, we first present the results of the successful application of transfer learning to one set of training and validation experiments. To demonstrate that these results are not serendipitous, we then carry out comprehensive cross-validation first using experimental, and then synthetic data. Finally, in Section \ref{sec:simple} we demonstrate that, unlike transfer learning, a linear calibration of the simulation outputs is not able to improve the predictions of the initial surrogate model.

Important contributions of this paper include:
\begin{description}
  \item[$\bullet$] Designing a method to include multi-modal experimental output data in model calibration, which has been traditionally carried out using only scalar output data.
  \item[$\bullet$] Implementing transfer learning in a complex forward-model$<$-$>$autoencoder architecture for a regression problem and, with no methods available to compare it against, also designing a simple calibration method as a baseline for transfer learning.
  \item[$\bullet$] Despite the extremely sparse data and the lack of benchmark datasets with multi-modal outputs, carrying out a very comprehensive cross-validation, including exhaustive cross-validation with the experimental data, as well as designing a large synthetic dataset mimicking our experiments, and validating our transfer learning method on these data.
\end{description}

\section{Data}
\label{sec:data}

\begin{table*}
\caption{Experimental and simulation inputs $x$.}
  \label{table:inputs}
  \begin{center}
    \scriptsize {
      \begin{tabular}{|lllllllll|}
        \hline
        \textbf{Input parameter}          & \multicolumn{1}{|l|}{ID} & \multicolumn{3}{l|}{Min simulated} & \multicolumn{3}{l|}{Max simulated}\\ \hline
        {Scale}                            & \multicolumn{1}{|l|}{1} & \multicolumn{3}{l|}{0.8} &   \multicolumn{3}{l|}{1.6}\\
        {Drive asymmetry mode 1,0}         & \multicolumn{1}{|l|}{2} & \multicolumn{3}{l|}{0} &   \multicolumn{3}{l|}{1\%}\\
        {Drive asymmetry mode 2,0, time 1} & \multicolumn{1}{|l|}{3} & \multicolumn{3}{l|}{-6\%} &   \multicolumn{3}{l|}{0}\\
        {Drive asymmetry mode 2,0, time 2} & \multicolumn{1}{|l|}{4} & \multicolumn{3}{l|}{-6\%} &   \multicolumn{3}{l|}{6\%}\\
        {Drive trough adjustment}         & \multicolumn{1}{|l|}{5} & \multicolumn{3}{l|}{-0.2} &   \multicolumn{3}{l|}{0.5}\\
        {Drive power adjustment}          & \multicolumn{1}{|l|}{6} & \multicolumn{3}{l|}{-0.25} &   \multicolumn{3}{l|}{0.5}\\
        {Drive energy adjustment}         & \multicolumn{1}{|l|}{7} & \multicolumn{3}{l|}{-0.25} &   \multicolumn{3}{l|}{0.5}\\
        {Preheat}                          & \multicolumn{1}{|l|}{8} & \multicolumn{3}{l|}{0} &   \multicolumn{3}{l|}{50}\\
        {Dopant fraction}                  & \multicolumn{1}{|l|}{9} & \multicolumn{3}{l|}{0.1\%} &   \multicolumn{3}{l|}{0.35\%}\\ \hline
      \end{tabular}
    }
  \end{center}
\end{table*}

The multi-modal experimental data consists of 10 ICF shots acquired during a so-called ``Bigfoot" campaign~\cite{bigfoot} at NIF.  The nine input parameters, defining our design space, are listed in Table \ref{table:inputs}. Six of them represent how the laser energy, converted into the X-rays, compresses the target capsule. These parameters include the energy, power, as well as the geometrical asymmetry parameterized in terms of two spherical-harmonic modes at two different times. The remaining inputs are related to the hydrodynamic scaling, additional energy deposited in the fuel known as the preheat, and the fraction of the dopant in the outer portion of the capsule.

Within the scope of this paper, we consider the experimental inputs to be known, and of the same type as the simulated inputs. In reality, however, some of the experimental inputs listed in Table \ref{table:inputs} had to be statistically inferred using the method described in~\cite{jimpop}, which adds uncertainty to any predictions we make.


Using the 10 Bigfoot experiments, our objective is to build a model that can predict the outcome of future shots, assuming that the variations in the design of the old and future shots are captured by the nine input parameters. Given that the design space is nine-dimensional, a robust predictive model cannot be constructed from 10 experiments alone. To sample the design space more adequately, we used a database of over 92 thousand simulations of the target capsule.
 The simulations are deterministic, employ a two-dimensional radiation hydrodynamic code HYDRA~\cite{hydra}, and took tens of millions of CPU hours~\cite{nora}~\cite{jimpop} to complete.

\begin{table}
\caption{Experimental and simulation outputs $y$.}
  \label{table:outputs}
  \begin{center}
    \scriptsize {
      \begin{tabular}{|llllll|}
        \hline
        \textbf{Output names} & \multicolumn{1}{|l}{} & \multicolumn{3}{l|}{Description} \\ \hline
        \textbf{\hspace{4pt} Scalars}                  & \multicolumn{1}{|c|}{ID} & \multicolumn{3}{l|}{} \\ \hline
        {\hspace{4pt} BT\_GRH}                         & \multicolumn{1}{|c|}{1} & \multicolumn{3}{l|}{Neutron bang time} \\
        {\hspace{4pt} BT\_SPIDER}                      & \multicolumn{1}{|c|}{2} & \multicolumn{3}{l|}{X-ray bang time} \\
        {\hspace{4pt} DSR\_AV}                         & \multicolumn{1}{|c|}{3} & \multicolumn{3}{l|}{Downscattered ratio} \\
        {\hspace{4pt} DT\_TION\_AV}                    & \multicolumn{1}{|c|}{4} & \multicolumn{3}{l|}{Temperature} \\
        {\hspace{4pt} P0\_HGXD\_090-078\_TI}           & \multicolumn{1}{|c|}{5} & \multicolumn{3}{l|}{Hot spot radius} \\
        {\hspace{4pt} DT\_VEL\_NTOF\_161-056}          & \multicolumn{1}{|c|}{6} & \multicolumn{3}{l|}{Velocity} \\
        {\hspace{4pt} LOG10\_XRAY\_YIELD\_22KEV}       & \multicolumn{1}{|c|}{7} & \multicolumn{3}{l|}{X-ray emission} \\
        {\hspace{4pt} LOG10\_DT\_YIELD\_AV}            & \multicolumn{1}{|c|}{8} & \multicolumn{3}{l|}{Neutron yield} \\
        {\hspace{4pt} BW\_GRH}                         & \multicolumn{1}{|c|}{9} & \multicolumn{3}{l|}{Neutron burn width} \\
        {\hspace{4pt} BW\_SPIDER}                      & \multicolumn{1}{|c|}{10} & \multicolumn{3}{l|}{X-ray burn width} \\ \hline
        \textbf{\hspace{4pt} Images}                   & \multicolumn{1}{|c}{} & \multicolumn{3}{l|}{X-ray image} \\ \hline
      \end{tabular}
    }
  \end{center}
\end{table}

The experimental and simulation outputs are listed in Table \ref{table:outputs}. The outputs include 10 scalar measurements describing how the target capsule responds to compression. These scalars include the time when the neutron and X-ray emission peaks (bang times), temperature, velocity, as well as the X-ray emission, and neutron yield - one of the most important ICF performance metrics. Each measured scalar has an associated error bar; these errors will be compared against prediction errors in Section \ref{sec:results}. In addition to the scalars, each output sample contains one, 60-by-60-pixel, single-channel image of the intensity of the X-ray emission. The image captures the central part of the target capsule known as the hot spot.

Some processing and data selection \cite{jimpop} has been applied to bring the experimental and simulated outputs into consistency but we recognize that the consistency is not perfect, especially for the images. There are multiple reasons contributing to this problem. First, half of our ICF shots had their X-ray images created using a different type of the imaging system than the other half.

Second, because each three-dimensional simulation would require on the order of one million CPU hours~\cite{clarkcpu}, our $\sim$92K simulations are only two-dimensional. Experimental variations in the unmodeled dimension are therefore visible in experimental images but not in the simulated images.

Third, running ensembles of ICF simulations and building surrogate models is fairly novel~\cite{nora} and we attempt to predict experimental images pixel-by-pixel for a suite of shots for the very first time. Important pieces of information about the images that are crucial for our work are not always available. For example, although image intensity is physically related to some of the scalar outputs, at present, we have no way of relating the experimental image intensities to the simulations; hence we simply divided all image amplitudes by their means before building predictive models.

Given all the aforementioned challenges and the fact that we work with only 10 experiments, we hope to calibrate only the overall structure of the hot spot in the X-ray images.


\section{Method}
\label{sec:method}

Building our calibrated predictive model consists of two steps. First, an initial surrogate model is trained on simulated data, as described in Section \ref{sec:initial_model}. Then, the surrogate model is exposed to the experimental data and retrained, as described in Section \ref{sec:transfer_learning}.

\subsection{Initial model}
\label{sec:initial_model}

To build the initial surrogate model, we closely follow the approach described in~\cite{rushilpnas}. Our goal is to build a surrogate model that predicts the ICF simulation outputs from the inputs residing within the ranges listed in Table \ref{table:inputs}. It is well known that, in the case of the ICF simulations, the mapping of the inputs $x$ to the outputs $y$ is nonlinear and that neural networks have been able to make more accurate predictions of the ICF outputs than simpler tree-based~\cite{rushilpnas} or linear~\cite{kustowski_speculative} models.

\begin{figure}
  \begin{center}
  \includegraphics[width=0.475\textwidth]{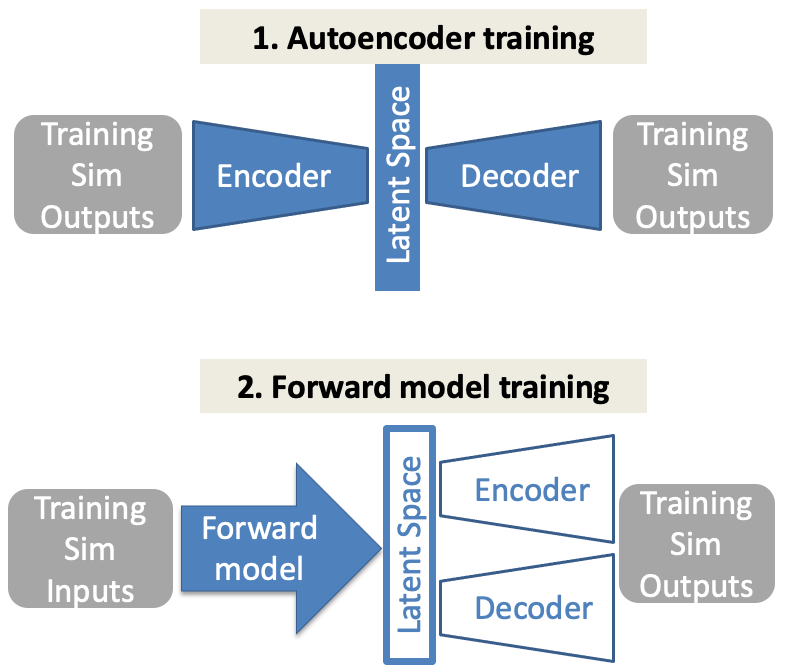}
  \caption{Architecture of the initial, surrogate model of the ICF simulations is the same as in \cite{rushilpnas}. Components of the architecture that are being trained at a given step are filled with color in the diagram. In the first step, the autoencoder is trained to compress all modalities of the simulated outputs, that is, the 10 scalars and one image. The compressed space consists of 32 latent variables $z$. Then, the forward model $F$ is trained to predict the latent variables, or, in other words, the encoded outputs. Once the entire architecture has been trained, predictions of the outputs $y$ can be made for new inputs $x$ by first applying the forward model to $x$, and then the decoder to the latent variables $z$.}
  \label{fig:initial_model}
  \end{center}
\end{figure}

\begin{table}
\caption{The degree of correlation $R^2$ between the simulated outputs and predictions of the initial surrogate model for validation samples.}
  \label{table:r2}
  \begin{center}
    \scriptsize {
      \begin{tabular}{|lll|}
        \hline
        \textbf{Output name}                           & \multicolumn{1}{|c|}{Autoencoder} & \multicolumn{1}{c|}{Forward} \\
        \textbf{}                                      & \multicolumn{1}{|c|}{} & \multicolumn{1}{c|}{model} \\
        \textbf{}                                      & \multicolumn{1}{|c|}{$R^2$ for} & \multicolumn{1}{c|}{$R^2$ for} \\
        \textbf{}                                      & \multicolumn{1}{|c|}{$D(E(y))$} & \multicolumn{1}{c|}{$D(F(x))$} \\ \hline
        \textbf{\hspace{2pt} Scalars}                  & \multicolumn{1}{|c|}{} & \multicolumn{1}{c|}{} \\ \hline
        {\hspace{2pt} BT\_GRD}                         & \multicolumn{1}{|c|}{1.000} & \multicolumn{1}{c|}{0.998} \\
        {\hspace{2pt} BT\_SPIDER}                      & \multicolumn{1}{|c|}{1.000} & \multicolumn{1}{c|}{0.999} \\
        {\hspace{2pt} DRS\_AV}                         & \multicolumn{1}{|c|}{1.000} & \multicolumn{1}{c|}{0.992} \\
        {\hspace{2pt} DT\_TION\_AV}                    & \multicolumn{1}{|c|}{0.999} & \multicolumn{1}{c|}{0.974} \\
        {\hspace{2pt} P0\_HGXD\_090-078\_TI}           & \multicolumn{1}{|c|}{1.000} & \multicolumn{1}{c|}{0.972} \\
        {\hspace{2pt} DT\_VEL\_NTOF-SPEC\_161-056}     & \multicolumn{1}{|c|}{1.000} & \multicolumn{1}{c|}{0.983} \\
        {\hspace{2pt} LOG10\_XRAY\_YIELD\_22KEV} & \multicolumn{1}{|c|}{1.000} & \multicolumn{1}{c|}{0.988} \\
        {\hspace{2pt} LOG10\_DT\_YIELD\_AV}            & \multicolumn{1}{|c|}{1.000} & \multicolumn{1}{c|}{0.990} \\
        {\hspace{2pt} BW\_GRH}                         & \multicolumn{1}{|c|}{1.000} & \multicolumn{1}{c|}{0.881} \\
        {\hspace{2pt} BW\_SPIDER}                      & \multicolumn{1}{|c|}{1.000} & \multicolumn{1}{c|}{0.932} \\ \hline
        \textbf{\hspace{2pt} Images}                   & \multicolumn{1}{|c}{1.000} & \multicolumn{1}{c|}{0.997} \\ \hline
      \end{tabular}
    }
  \end{center}
\end{table}

Predicting the outputs directly from the inputs might be possible, however,~\cite{rushilpnas} advocate that physical relationships within data are better preserved if the outputs are first compressed by taking advantage of the correlations within the data. Since these correlations are not linear (with the exception of the two measurements of the bang time),~\cite{rushilpnas} compress the outputs using a nonlinear method, a Wasserstein autoencoder~\cite{wasserstein}. The autoencoder consists of two neural networks: an encoder $E$, which compresses the outputs $y$ into a set of latent variables $z$, and a decoder $D$, which reconstructs $y$ from $z$ (Figure \ref{fig:initial_model}). The innermost layers of the encoder and decoder and fully connected, while the outer layers consist of convolutional layers for predicting images and a fully connected branch for predicting scalars. Because of the correlations within $y$, as well as a fairly simple structure of the X-ray image, 32 latent variables are sufficient to reconstruct all simulated outputs nearly perfectly, with the validation $R^2$ exceeding 0.999 (Table \ref{table:r2}). The standard coefficient of determination $R^2$ is used here to evaluate the  the model predictions, where $R^2=1$ indicates a perfect correlation between the simulated and predicted outputs. For images, $R^2$ was calculated on a pixel-by-pixel basis.

Once the outputs have been compressed into a reduced set of latent variables, an inverse model $I$ and a forward model $F$ are trained to find the mapping between the inputs and the latent variables. The action of each of the four models is summarized in the following set of equations:

\begin{equation} \label{eq:initial_model}
z = E(y), y = D(z), z = F(x), \ x = I(z).
\end{equation}

The forward and inverse models are neural networks with fully-connected layers and special cyclic-consistency regularization developed in~\cite{rushilpnas}. For simplicity, Figure \ref{fig:initial_model} shows only the forward model; the inverse model plays an important role in the development of the initial model but not in transfer learning.

The validation $R^2$ for all scalar outputs predicted from the inputs by

\begin{equation} \label{eq:forward}
y = S(x) = D(F(x))
\end{equation}

is very high (Table \ref{table:r2}). This confirms that the number of training simulations is adequate for the nine-dimensional design space, and that the response surface is sufficiently smooth to be represented by $\sim$3.5 simulations per dimension.

\subsection{How to apply transfer learning in this new architecture?}
\label{sec:transfer_learning}

\begin{figure*}
  \begin{center}
  \includegraphics[width=0.95\textwidth]{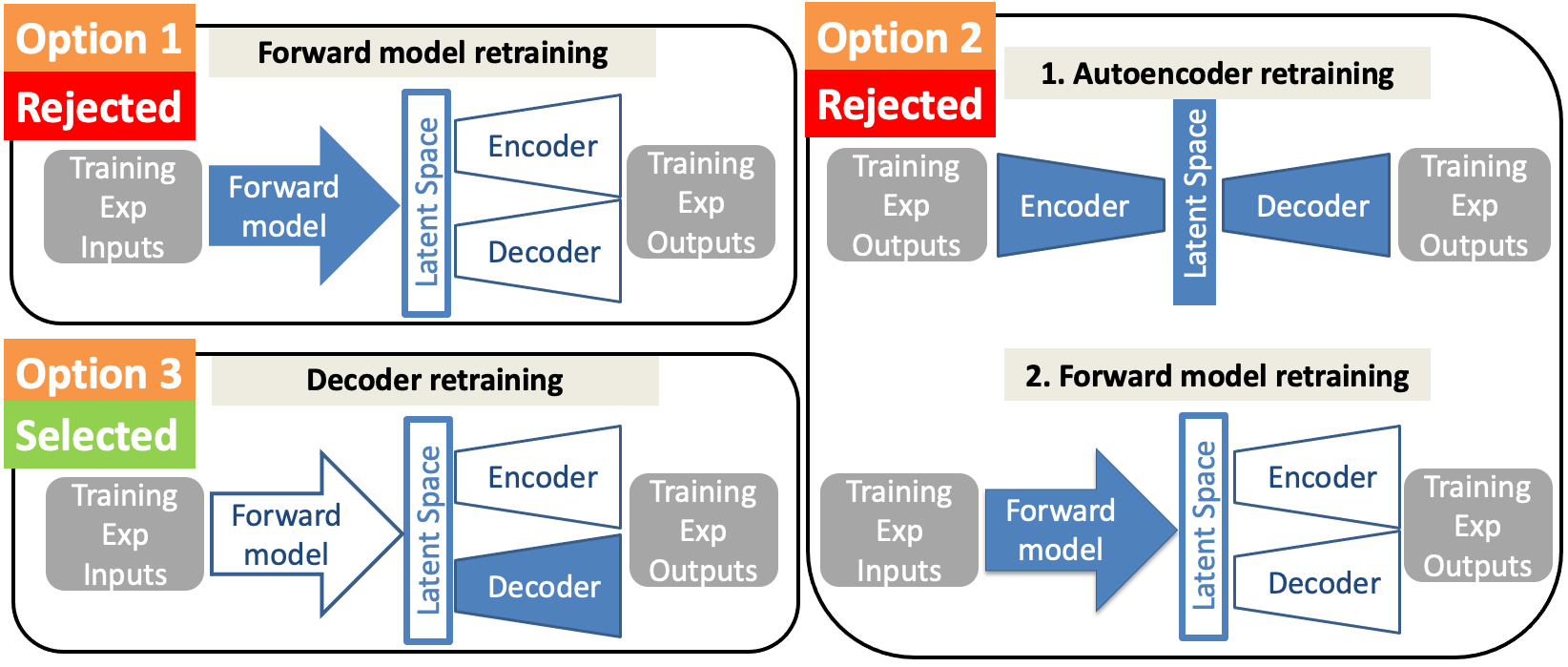}
  \caption{Transfer learning options considered in this paper. Components of the architecture that are being trained are filled with color. In the first option, the autoencoder is fixed and only the forward model is being retrained. The second option involves retraining of the autoencoder, and then of the forward model. In the preferred, third option, the forward model and encoder are fixed and only the decoder is retrained.}
  \label{fig:transfer_learning_options}
  \end{center}
\end{figure*}

Once the initial model has been trained on simulated data, we wish to retrain it to match the experimental data. In contrast to standard applications discussed in the machine learning literature, where transfer learned models merely classify images, our retrained model has a more challenging task of generating images and scalars as outputs. With only 10 experimental shots available, we expect our algorithm to suppress only a fairly simple, systematic bias of the simulations relative to the experiments. We also assume that the regularizing effect of the initial, simulation-trained model is sufficiently strong to enable robust refinement of thousands of weights in the neural network.

In the standard transfer learning procedure described in the literature~\cite{yosinski}, typically the last layer, or layers, within a single neural network get retrained to match the new data. Because our initial model is not a single neural network, it is not obvious which part of our system of networks needs to be retrained. Since there is no formal guidance in the literature for it, we implemented and tested three quite different strategies. Presenting numerical results of these tests would be worth a separate, method-oriented publication, and is beyond the scope of this paper. We now discuss, however, these three options since they make different assumptions about the simulation bias and imply a different role of transfer learning in removing it.

In the first option, shown in Figure \ref{fig:transfer_learning_options}, we retrained only the last layer of the forward model. This strategy has a clear interpretation. By keeping the autoencoded outputs fixed, we assumed that the latent representation of the outputs, derived from the simulations, is perfectly suitable for the experimental data. The simulation bias is therefore defined in this case as an incorrect position on the latent space, which can be corrected by the retraining of the forward model. This further implies that the bias is caused by the inadequacy of the simulation rather than the systematic discrepancy between simulated and experimental outputs that could be caused, for example, by the simulation post-processing procedure. Using this method, however, we were able to make only very small corrections without inflating the generalization error. This suggests that the correlations in the outputs extracted by the autoencoder from the simulation outputs are not exactly the same as the correlations in the experimental data.

To allow for corrections in the autoencoder, in the second option, we retrained the innermost layers in the encoder and in the decoder. This allowed for a good autoencoder reconstruction of the experimental outputs but it also made the autoencoder incompatible with the initial forward model, because the initial forward model predicts the outputs compressed by the initial autoencoder. We therefore subsequently retrained the last layer of the forward model as well. Once again, we found that only a small correction was possible without inflating the generalization error. This suggests that the bias exists not only in the correlations in the data extracted by the autoencoder but also in the forward model, which emulates the simulation. Furthermore, these biases cannot be corrected  independently of each other.  In order to correct them simultaneously, we tested the third option.

In the third option, we kept the encoder and forward model fixed and retrained only the innermost layer of the decoder. This fully connected layer is shared by all data modalities. Two subnetworks branch out from this innermost layer: a second fully connected layer for predicting the scalars, and a set of convolutional layers for predicting the image. Retraining of the decoder can therefore correct the imperfect correlations represented by the initial autoencoder. Because we retrain the innermost layer of the decoder, different data modalities are simultaneously affected by this change rather than being retrained independently of each other.

The third option gave us by far the best generalization error while making appreciably large corrections, and we will discuss only this option thereafter. The key step in the method is the retraining of the decoder. Relative to the training of the initial autoencoder, we added a standard L2 regularization of the model weights at a 5\% level, decreased the number of training iterations from 200,000 to 100, and decreased the learning rate from 1e-4 to 3e-5. Such small number of iterations is sufficient because the initial model is already fully trained on simulations and it only needs to be slightly modified to reduce the bias. Using a small number of iterations also prevents overfitting to the sparse experimental data and catastrophic forgetting \cite{catastrophic} of what the model has learned from the simulations. The optimization objective in transfer learning can be written as 

\begin{eqnarray} 
    \min_{\bm{\theta}}  (\| \mathbf{y}_{img} - \mathbf{\hat{y}}_{img} \|_2^2 \
    + \gamma_{sca} \| \frac{\mathbf{y}_{sca} - \mathbf{\hat{y}}_{sca}}{\bm{\sigma}_{y_{sca}}} \|_2^2 \nonumber \\
    + \lambda_{reg} \| \bm{\theta} \|_2^2), \label{eq:objective} 
\end{eqnarray}

where the weight $\gamma_{sca}$ controls how well the scalar outputs $\bm{y}_{sca}$ are matched relative to the image outputs $\bm{y}_{img}$, the hat symbol indicates model predictions, and $\lambda_{reg}$ controls the importance of the regularization imposed on the neural network weights $\bm{\theta}$.

In Section \ref{sec:simple}, we used only the L2 norm of the prediction error as the optimization metric but in Section \ref{sec:results}, we divided the error for scalar outputs by the measurement error $\bm{\sigma}_{y_{sca}}$ thus using a $\chi^2$ to measure the goodness of fit. A standard min-max normalization was applied to $\bm{y_{sca}}$ before training but it was removed in the computation of the loss function whenever the prediction error was divided by $\bm{\sigma}_{y_{sca}}$. Pixel-wise measurement errors have not been estimated for images. Images were normalized by simply dividing intensities by the mean value of each image because, at this point, we have no way of relating the experimental and simulated image intensities in a reliable way, as mentioned in Section \ref{sec:data}.
 
The number of gradient descent iterations, the learning rate, as well as the weights $\gamma_{sca}$ and $\lambda_{reg}$, were determined through cross-validation described in Section \ref{sec:xval_data}. Once the retrained decoder $D^{TL}$ has been computed, the predictions of the outputs are made by

\begin{equation} \label{eq:tlpred}
y = S^{TL}(x) = D^{TL}(F(x)),
\end{equation}

where $F$ is the simulation-trained forward model.


\section{Results}
\label{sec:results}

\begin{figure*}
  \begin{center}
  \includegraphics[width=0.95\textwidth]{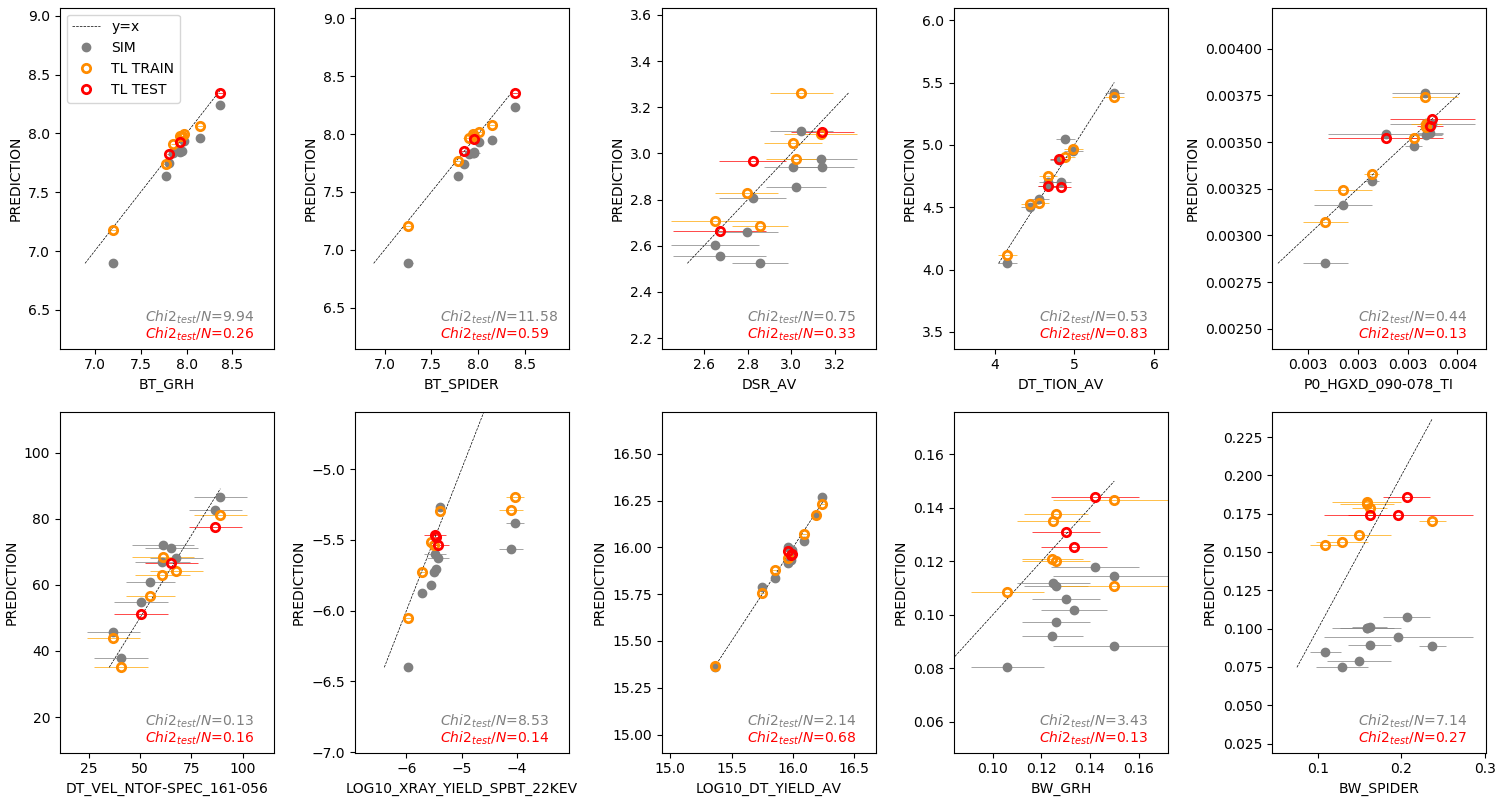}
  \caption{Transfer learning results for the 10 scalar outputs using seven shots for retraining and three shots as validation. The horizontal axis corresponds to the observed values and the horizontal bars show the experimental error. The vertical axis corresponds to the predicted values. Gray dots are the predictions of the simulation-trained model. Color dots are the predictions of the transfer learned model for the seven training shots (orange) and three validation shots (red). The $\chi^2/N$ metric is computed only for the validation data  and is divided by the number of samples $N$ so that the error larger than one indicates that the prediction error is larger than the experimental error. }
  \label{fig:scalar_improvement}
  \end{center}
\end{figure*}

\begin{figure*}
  \begin{center}
  \includegraphics[trim=140 80 360 70, clip, width=0.65\textwidth]{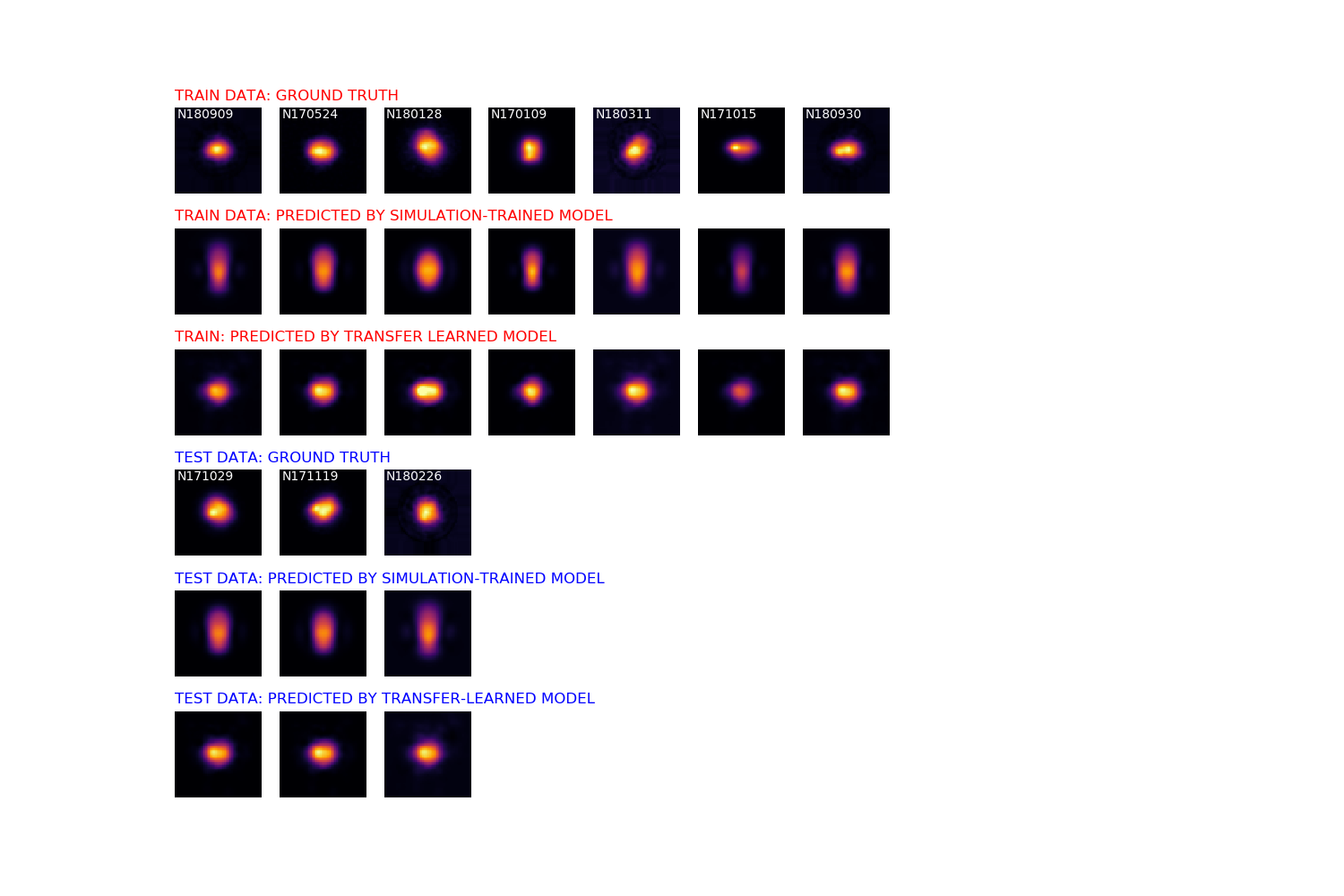}
  \caption{Transfer learning results for the X-ray images. The top three rows correspond to the  seven training shots and the bottom three rows correspond to the three validation shots. Each image represents the X-ray emission at the central part of the target capsule - the hot spot. For each experiment, the predicted images are plotted using the same colormap range as the corresponding ground truth image. The name of each ICF shot is printed on top of the ground truth images.}
  \label{fig:image_improvement}
  \end{center}
\end{figure*}

Our first test consists of randomly selecting seven ICF shots to retrain the initial model, while the remaining three shots are used to evaluate the generalization error. Figure \ref{fig:scalar_improvement} compares the predictions of the initial (Equation \ref{eq:forward}) and calibrated (Equation \ref{eq:tlpred}) models against the observed values for the 10 scalar outputs. Scalars that experienced the largest correction are the two burn widths (BW). Calibration  brought them to within the experimental error from the ground truth values for the majority of the shots. The initial predictions of both bang times (BT) are well correlated with the observations but are also systematically underestimated, which, in combination with the small observational error, lead to the large values of $\chi^2/N$. Calibration dramatically reduced this error. Large improvement is also observed for both YIELDs. For the remaining five scalars, the effect of calibration is less dramatic and requires more thorough validation.

Simultaneously with improving predictions of the scalar outputs, transfer learning improved  predictions of the X-ray images (Figure \ref{fig:image_improvement}). Experimental images display a range of shapes from slightly prolate to very oblate hot spots. The initial model systematically predicted very prolate hot spots and calibration correctly modified the shapes to more round or slightly oblate. Transfer learning also preserved minor variations between the shots, such as the lower intensity in N171015, and the slightly prolate shape in N170109. Hot spots in  N180128 and N180311 are elongated in the diagonal direction; a feature that was not modeled in our two-dimensional simulations, hence surrogate models cannot reconstruct it.

Predicting whether the hotspot is prolate or oblate is the main characteristic we expected to reconstruct in images. Although transfer learning improved the predicted shapes, there is not a lot of variability in the initial and calibrated model predictions. The capability of transfer learning to handle more diverse shapes will be further demonstrated in Section \ref{sec:xval_synth}.

In summary, our model calibration has clearly improved the predictions of the initial model but validation with only three experiments could be serendipitous. In order to demonstrate that the improvement is systematic, we now turn to a more thorough cross-validation using experimental (Section \ref{sec:xval_data}) and synthetic data (Section \ref{sec:xval_synth}).


\subsection{Cross-validation using experimental data}
\label{sec:xval_data}

\begin{figure*}
  \begin{center}
  \includegraphics[width=0.975\textwidth]{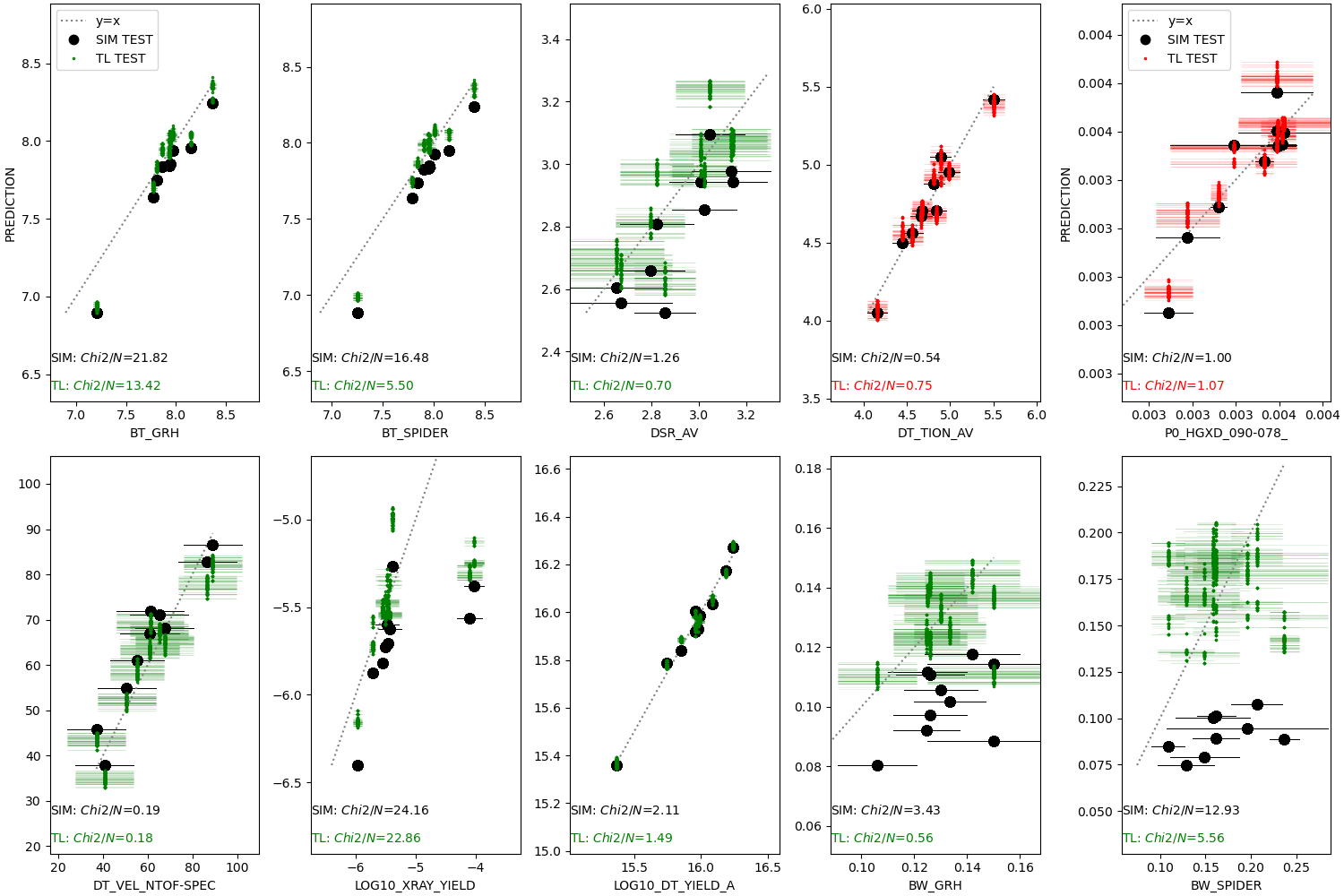}
  \caption{Cross-validation with 500 train-test splits. The horizontal axis corresponds to the observed values of the scalar outputs and the vertical axis corresponds to the predicted values. Horizontal bars are the measurement errors. Each experiment is predicted only once by the initial model (black dots) and multiple times by transfer learned models for different train-test splits (color dots). The green color indicates that calibration improved the predictions over the initial model in terms of $\chi^2/N$, and the red color indicates that it did not. Dotted line is the $y=x$ line plotted for reference. $\chi^2/N$ was calculated after concatenating the three validation samples from all train-test splits.}
  \label{fig:xval_data}
  \end{center}
\end{figure*}

\begin{figure}
  \begin{center}
  \includegraphics[trim=50 10 60 20, clip, width=0.475\textwidth]{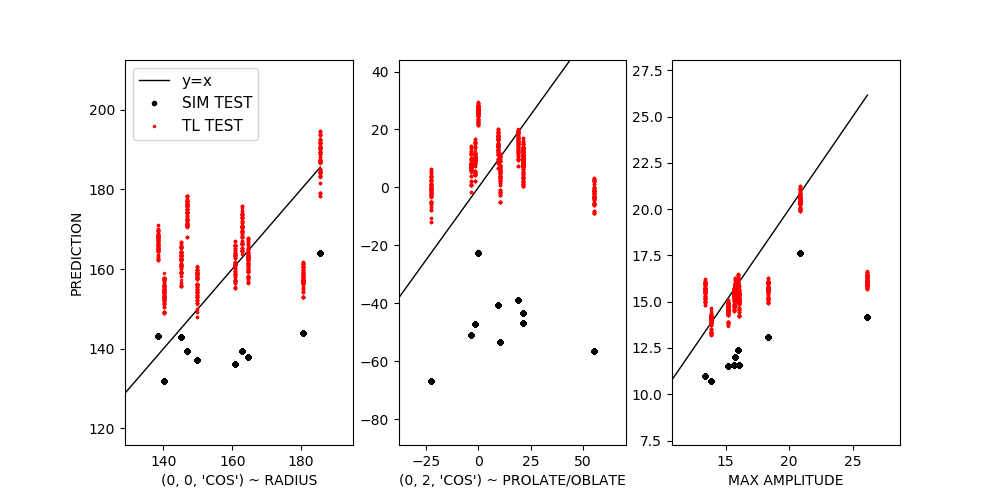}
  \caption{Cross-validation with 500 train-test splits. X-ray images are represented by three parameters: radius, shape, and maximum amplitude. The horizontal axis corresponds to the observed values and the vertical axis corresponds to the predicted values. Each experiment is predicted only once by the initial model (black dots) and multiple times by transfer learned models corresponding to different train-test splits (red dots).
     Those familiar with the Legendre polynomial expansion of round objects should note that the Gauss-Laguerre mode (0, 2, COS) has opposite sign to the Legendre mode 2.}
  \label{fig:xval_data_images_lg}
  \end{center}
\end{figure}

Comprehensive cross-validation is essential for understanding whether a model can systematically make accurate predictions. With only 10 experiments, we could consider the exhaustive cross-validation since there are only 120 possible splits into seven training samples and three validation samples. For practical reasons, we chose a slightly different strategy; we randomly split the data 500 times with replacement. We consider this approach to be approximately equivalent to the exhaustive cross-validation.

For each of the 500 train-test splits, we started with the same initial model, retrained it using seven experiments, and evaluated the validation error on the remaining three samples. In other words, we evaluated 500 different calibrated models. For each scalar output, we computed the $\chi^2/N$ metric after concatenating the three samples from all 500 splits. Since we divided the $\chi^2$ by the number of validation samples $N$, we can interpret the values smaller than one as an indication that the prediction error is smaller than the experimental error.

The cross-validation confirmed that the calibration systematically improves the predictions of the initial model (Figure \ref{fig:xval_data}). The correction for both burn widths (BW), both bang times (BT), and DT\_YIELD is large and the drop in $\chi^2/N$ varies between a factor of two and six for these four scalars. The remaining scalars show smaller changes. Overall, transfer learning improved the $\chi^2/N$ for eight out of ten scalars. Calibration did not generate poorly predicted outliers for any of the 500 splits confirming the robustness of the method.

In order to summarize the cross-validation results for images, we parameterized each image in terms of three numbers (Figure \ref{fig:xval_data_images_lg}). The first two parameters come from the Gauss-Laguerre~\cite{laguerre},\cite{mikelaguerre} expansion: the (0, 0, COS) mode approximates the radius of the hot spot, and the (0, 2, COS) coefficient is positive for the oblate hots pots and negative for the prolate hot spots. The sign of this coefficient is  opposite to the Legendre mode 2 - a standard characteristics in the ICF community. The third parameter is the maximum amplitude in the image.

Calibration systematically improved predictions for all three image parameters suppressing the bulk shift between the simulation and experiments. All 500 predictions are clearly better then the predictions of the initial surrogate model with the exception of radius predictions of four shots with the smallest radius. It is also remarkable that none of the three image parameters was included as a metric in the optimization; transfer learning minimizes a pixel-wise L2 norm of the images. We also need to keep in mind that, as discussed Section \ref{sec:data}, the amplitudes are affected by dividing images by their mean values, which might have broken the physical consistency with the scalar outputs. These issues increase measurement uncertainties for the maximum amplitude as possibly also the radius, but we do not have quantitative estimates of these error bars.


\subsection{Designing synthetic data for additional cross-validation}
\label{sec:xval_synth}

\begin{figure*}
  \begin{center}
  \includegraphics[width=0.95\textwidth]{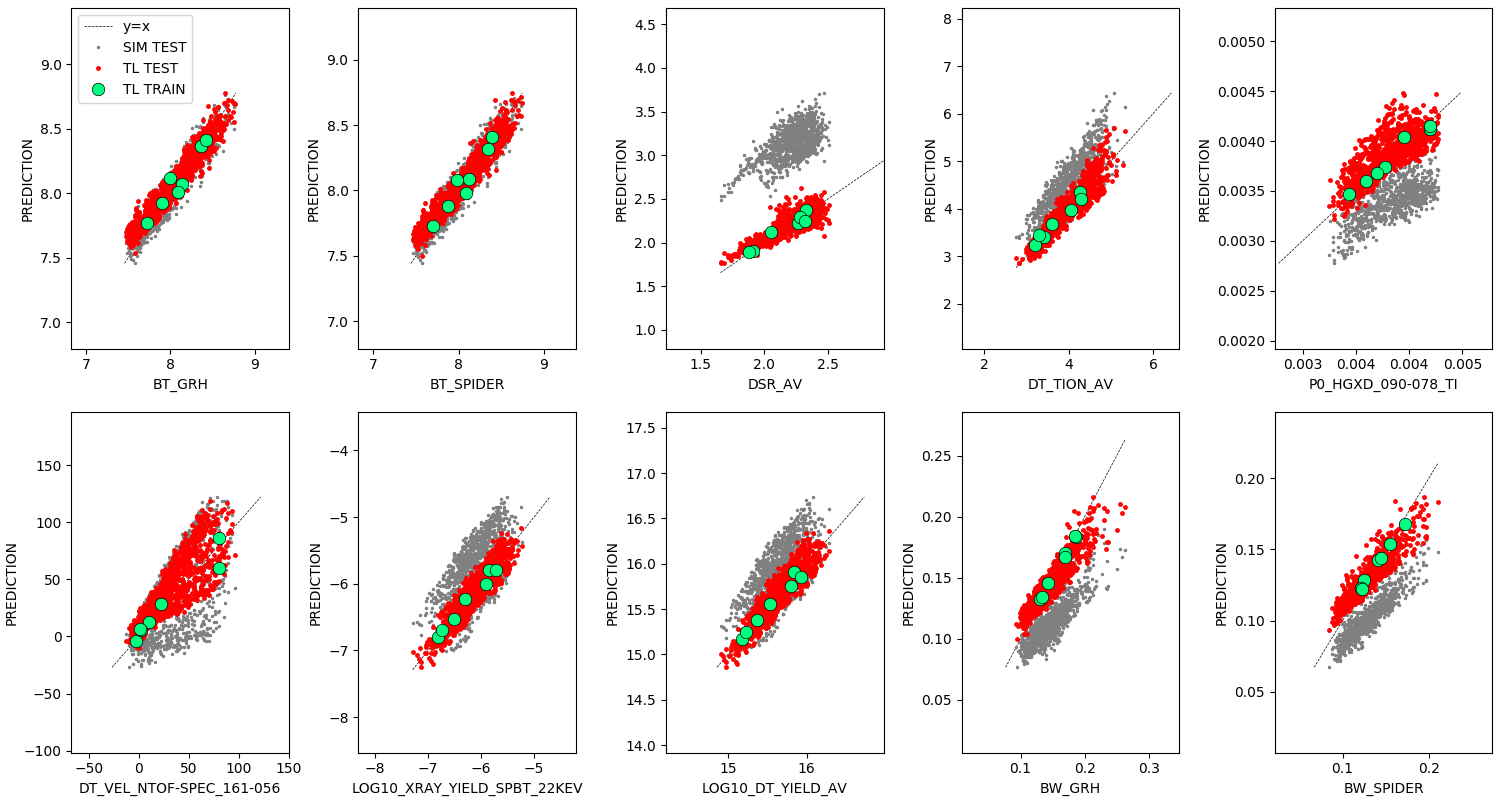}
  \caption{Transfer learning cross-validation with synthetic data. Horizontal axis corresponds to the values of the scalar outputs from the ``experiments" and the vertical axis corresponds to the predicted values. Green circles are the seven ``experimental" samples used for retraining. Dots show the 1,000 validation samples for the initial model (gray) and calibrated model (red).}
  \label{fig:xval_synth_scalars}
  \end{center}
\end{figure*}

\begin{figure*}
  \begin{center}
  \includegraphics[trim=140 80 110 70, clip, width=0.95\textwidth]{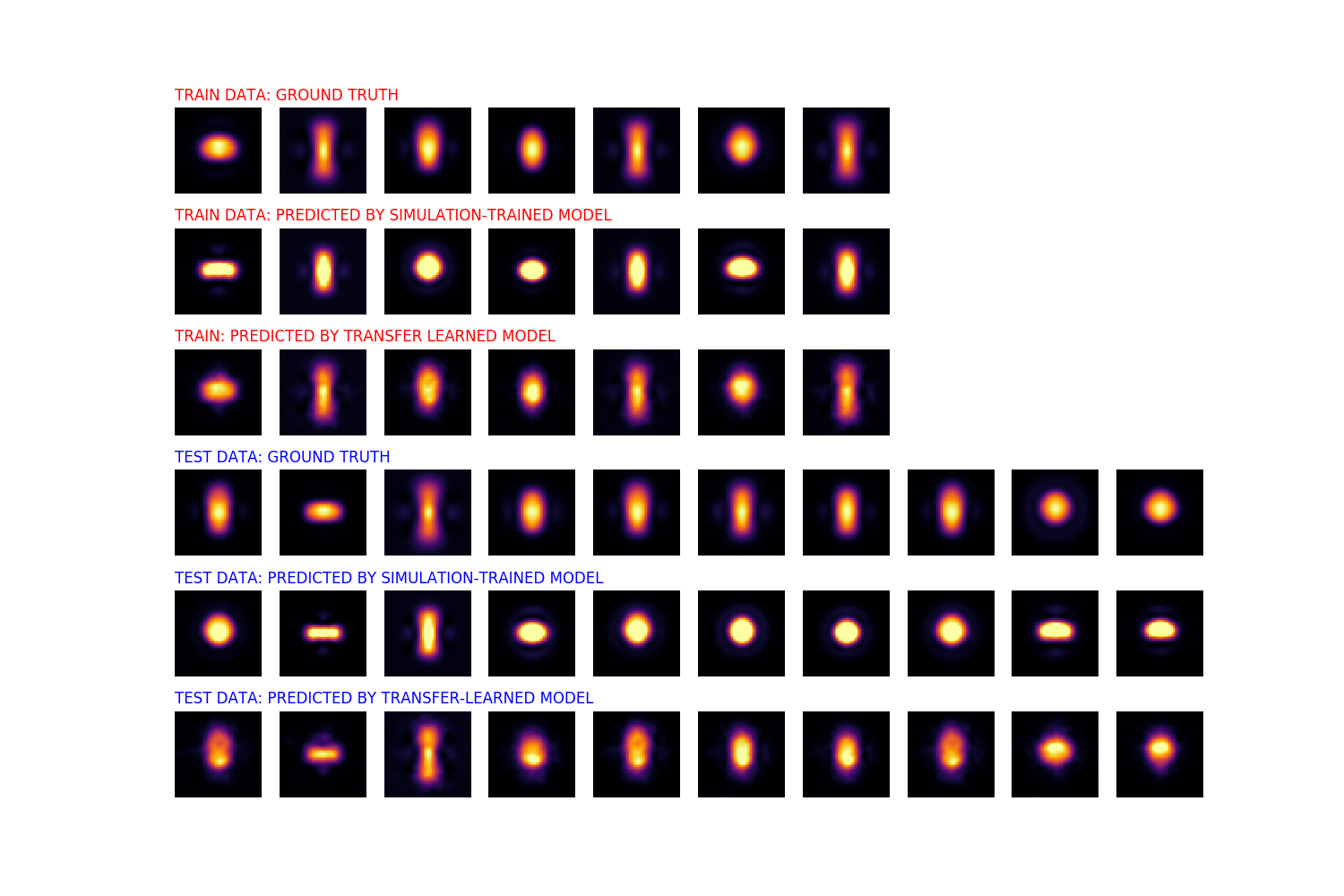}
  \caption{Transfer learning cross-validation with synthetic data for X-ray images. The top three rows show the seven ``experiments" used to retrain the model. The bottom three rows show 10 randomly selected validation shots. For each experiment, predicted images are plotted using the same colormap range as the corresponding ground truth image.}
  \label{fig:xval_synth_images}
  \end{center}
\end{figure*}

\begin{figure}
  \begin{center}
  \includegraphics[trim=50 10 60 20, clip, width=0.475\textwidth]{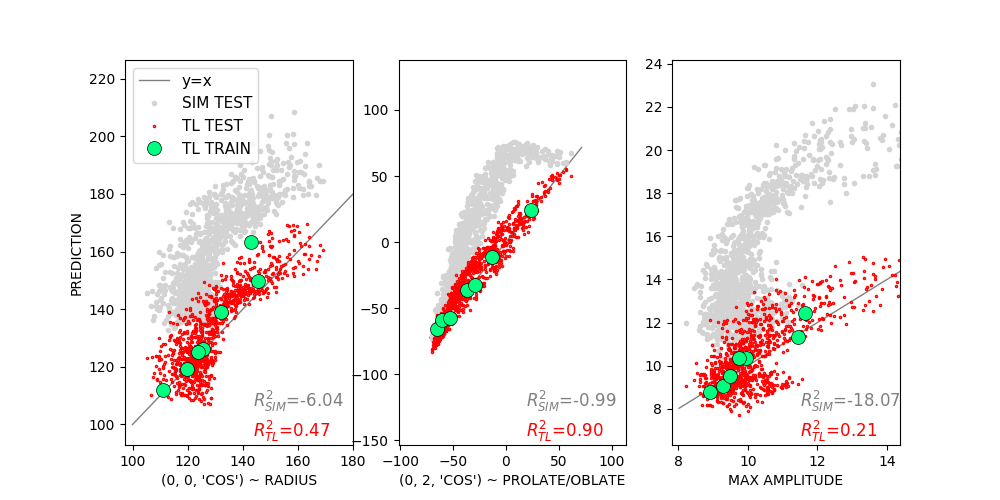}
  \caption{Transfer learning cross-validation with synthetic data for X-ray images. The images are represented by three parameters: radius, shape, and maximum amplitude. The horizontal axis shows the values from the ``experiments" and the vertical axis show the predicted values. Green circles are the seven "experimental" samples available for retraining. Dots show the validation samples for the initial model (gray) and calibrated model (red). Those familiar with the Legendre polynomial expansion of round objects should note that the Gauss-Laguerre model (0, 2, COS) has the opposite sign to Legendre mode 2.}
  \label{fig:xval_synth_images_lg}
  \end{center}
\end{figure}

After having completed the cross-validation using experimental data, we now turn to a more extensive cross-validation using synthetic experiments. With a thousand of validation samples available, we expect to better understand which aspects of the simulation bias can or cannot be corrected by transfer learning.

First, we need to decide how to generate the synthetic data to be representative of our problem. The authors of~\cite{kustowski2019} generated the synthetic ICF experiments by modifying certain physics parameters in the simulation. Then, they transfer learned from the initial surrogate, trained on simulations with the nominal physics, to the domain of the ``experiments" simulated by the perturbed physics. In that work, the authors run a very fast, but also very simplified, simulation~\cite{jag}~\cite{springer}, which cannot model real ICF experiments. Repeating this procedure with two-dimensional HYDRA simulations, would require at least 100,000 CPU hours to generate 1,000 samples~\cite{nora} and the process would likely need to be repeated multiple times to define reasonable physics perturbations. New simulations would also need additional quality assurance.

Motivated by these practical considerations, instead of running a new set of simulations, we  generated the synthetic data by employing our uncalibrated surrogate model of the simulations to make predictions.
First, in order to generate the nominal simulation data set, we fixed four of the nine inputs parameters to the values listed in Table \ref{table:xval_synth}. The other five input parameters were sampled randomly in the original ranges listed in Table \ref{table:inputs}. Having generated 92,000 inputs samples, we then predicted the outputs using the uncalibrated surrogate model and Equation \ref{eq:forward}. Then, to generate the synthetic experiments, we perturbed the values of the asymmetry and preheat parameters, kept them fixed, and repeated the procedure generating 1,000 ``experiments''. The preheat perturbation predominantly affects the scalars, while the asymmetry parameter perturbs the way  X-rays compress the target capsule thereby strongly affecting X-ray images. Furthermore, we fixed the values of scale and dopant fraction to 1.0 and 0.28\%, respectively, in both datasets. The experimental values of these two parameters vary in narrow ranges of 0.89-1.0 for the scale, and 0.21-0.28\% for the dopant fraction (with the exception of one shot). The simulated ranges of 0.8-1.6 and 0.1-0.35\%, respectively, are much wider and designed with the aim of making predictions rather than fitting experimental data. By suppressing variations in the scale parameter, we prevent transfer learning from calibrating against "experiments" at scales that are currently not attainable at NIF.

This procedure generated two lower-dimensional, and physically inconsistent data sets for transfer learning. We consider this approach to be nearly equivalent to running a new set of simulations since our surrogate model fits the simulations at the validation $R^2 > 0.932$ level for all but one outputs (Table \ref{table:r2}). We also recognize that the true simulation-experiment discrepancy could not be reproduced even if we run a new set of simulations with perturbed physics because the choice of the physics perturbation would be arbitrary.
The main objective of our cross-validation, however, is to determine which parts of the bias can or cannot be corrected by retraining with only seven shots, and our synthetic data are adequate for this task.

\begin{table}
\caption{Parameterization of the nominal and perturbed synthetic data used in cross-validation.}
  \label{table:xval_synth}
  \begin{center}
    \scriptsize {
      \begin{tabular}{|llllllll|}
        \hline
        \textbf{Input parameter}           & \multicolumn{3}{|c|}{Nominal } & \multicolumn{3}{c|}{Perturbed }\\
        \textbf{}                          & \multicolumn{3}{|c|}{model} & \multicolumn{3}{c|}{ value}\\ \hline
        {Drive asymmetry mode 2,0, time 1} & \multicolumn{3}{|l|}{0\%} & \multicolumn{3}{l|}{-5\%}\\
        {Preheat}                          & \multicolumn{3}{|l|}{5} & \multicolumn{3}{l|}{20}\\
        {Scale}                            & \multicolumn{3}{|l|}{1} & \multicolumn{3}{l|}{1}\\
        {Dopant fraction}                  & \multicolumn{3}{|l|}{0.28\%} & \multicolumn{3}{l|}{0.28\%}\\ \hline
      \end{tabular}
    }
  \end{center}
\end{table}

We now have the two five-dimensional data sets, one of which we treat as the actual simulations, and the other one as the ``experiments". To fit the simulation dataset, we trained a new, initial surrogate model using exactly the same procedure as in the case of the original, nine-dimensional data set described in Section \ref{sec:initial_model}. Figure \ref{fig:xval_synth_scalars} shows the results of transfer learning of this new surrogate model to match the ``experiments" from the other five-dimensional data set. Using only seven ``experiments" for retraining, calibration successfully removed the bulk shift of the simulations with respect to the ``experiments". There are no outliers among the calibrated predictions: every one of the 1,000 validation samples has moved approximately as much as its neighbors. It is also worth noting that the transfer learning correction worked not only in the region between the training samples (green circles) but everywhere inside the sampled region, where we run simulations. This indicates some extrapolation capability of the transfer learning method.

Other than the bulk shift, the simulation-trained predictions show appreciable scatter indicating that the response of each scalar to the perturbations in the asymmetry and preheat is complex and also depends on other inputs. This scatter could potentially be reduced by transfer learning had we (i) used many more than seven ``experiments" for retraining, (ii) retrained the entire decoder and forward model, and (iii) applied the same hyperparameters as those used during the initial surrogate training. Instead, to make our cross-validation relevant, in the calibration we employed an identical strategy as in transfer learning with real experiments: we used only seven ``experiments", retrained only the innermost layer of the decoder, and used regularization and only 100 iterations in the optimization. With such limited data, model capacity, and the number of training iterations, it is not surprising that transfer learning did not reduce the scatter in the predictions.

Calibration results for the images are shown in Figure \ref{fig:xval_synth_images}. In contrast to the real experiments, we now see more variability in both the ground truth and predicted hot spots. Calibration modified the simulated images towards more prolate shapes, consistently with the ground truth images. This shift was successfully applied to hot spots of all shapes: prolate, round, and oblate. However, not all calibrated images have a prolate shape; whenever the target image was oblate, the calibrated model correctly preserved the oblate shape of the hot spot.

To extend the analysis of the images to all 1,000 validation samples, once again, we parameterized each image in terms of the radius, shape, and maximum image amplitude (Figure \ref{fig:xval_synth_images_lg}). As before, we see that transfer learning systematically modified images towards more prolate shapes. For example, the initial model prediction with (0, 2, COS) = 25, indicating an oblate shape, changed after the calibration to about \mbox{-20}, indicating a prolate shape, consistently with the target ``experiment". Calibration of this mode was extremely  successful lifting the $R^2$ metric  from a negative value to 0.9. It is remarkable that using a pixel-wise L2 metric in the optimization, transfer learning was able to correct a non-pixel characteristics of an image with such great accuracy.

Systematic biases in the predictions of the radius and maximum image amplitude were also dramatically suppressed by the calibration, although the residual scatter appears to be slightly larger than in the scalar outputs. Reconstruction of the amplitudes may be hampered by the fact that we removed mean amplitudes from all images while image amplitudes are physically related to some scalars, as described in Section \ref{sec:data}. To compensate for this inconsistency, transfer learning might have changed the radius as well.

Overall, cross-validation with the synthetic data confirmed that using just seven ``experiments'', it is possible to retrain the surrogate and remove the bulk shift between simulations and experiments for both scalar and image data.


\section{Designing simple calibration as a baseline for transfer learning}
\label{sec:simple}

\begin{figure*}
  \begin{center}
  \includegraphics[width=0.95\textwidth]{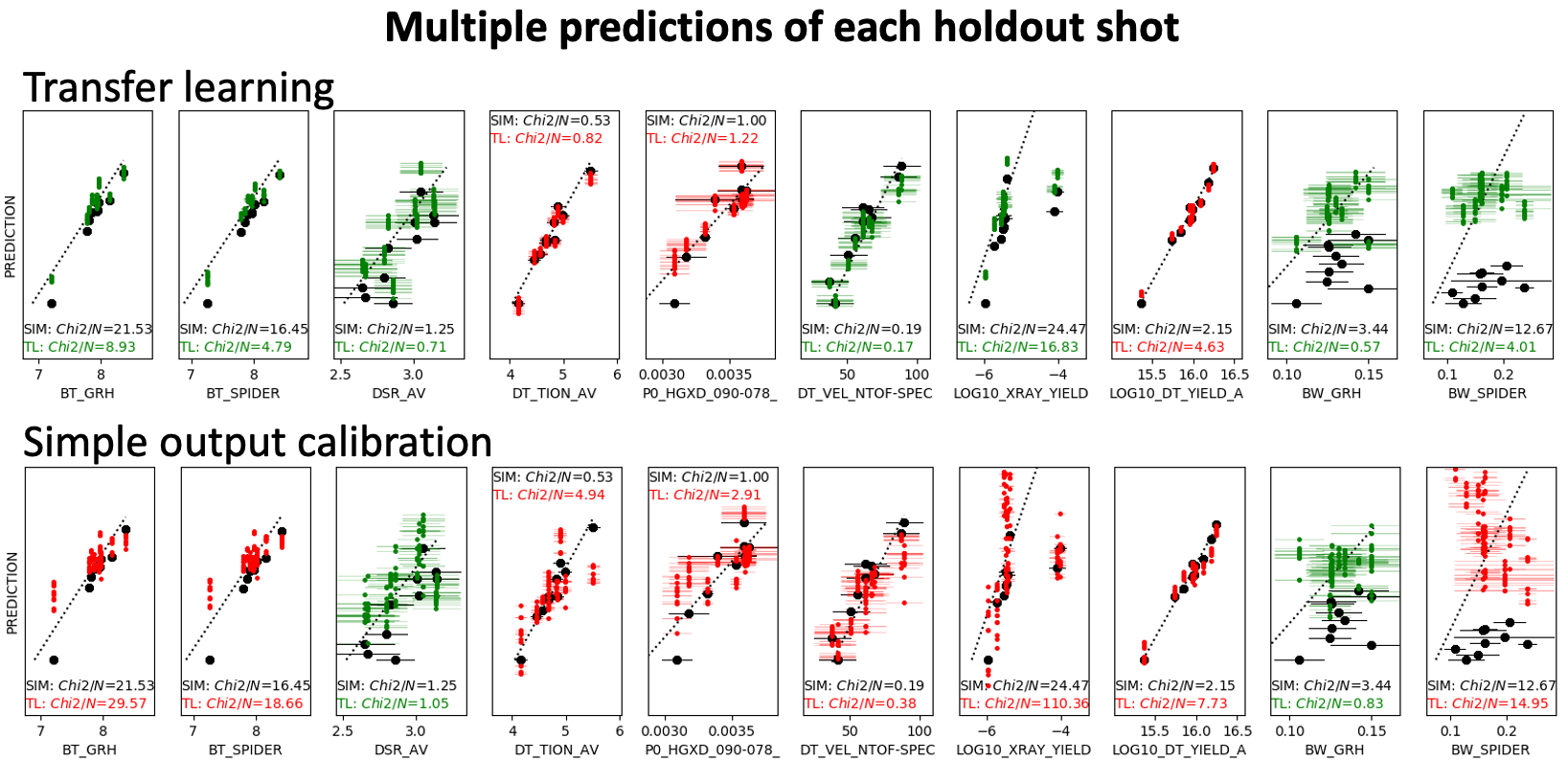}
  \caption{Comparison of the transfer learning (top row) with the simple, linear output calibration (bottom row). Each shot is hold out and predicted 15 times by seven, randomly selected shots. Only the holdout shots are shown. The horizontal axis corresponds to the experimental values. The vertical axis corresponds to the predicted values. Predictions of the simulation-trained model are shown in black and predictions of the calibrated model are shown in color. Green color indicates that the calibration has improved the $\chi^2/N$ metric and red color indicates that it did not. The $\chi^2/N$ metric is computed for a concatenated vector of the 15 predictions from all 10 holdout shots.}
  \label{fig:bagging_all}
  \end{center}
\end{figure*}

\begin{figure*}
  \begin{center}
  \includegraphics[width=0.95\textwidth]{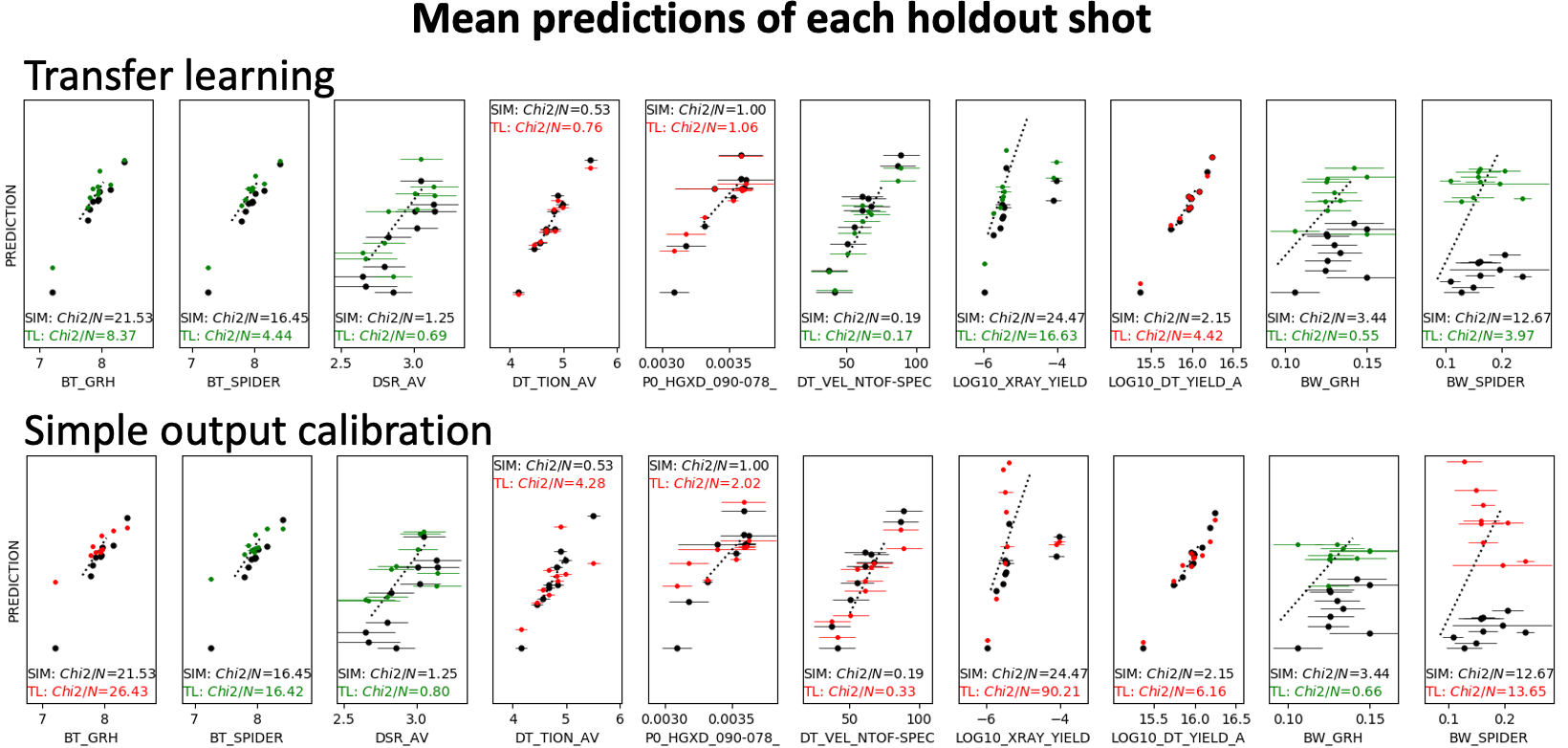}
  \caption{Bootstrap aggregation of the results from in Figure \ref{fig:bagging_all}. For each holdout shot, all predictions are averaged out. The $\chi^2/N$ metric is computed from the mean predictions over the 10 holdout shots.}
  \label{fig:bagging_means}
  \end{center}
\end{figure*}

\begin{figure}
  \begin{center}
  \includegraphics[trim=50 10 60 20, clip, width=0.475\textwidth]{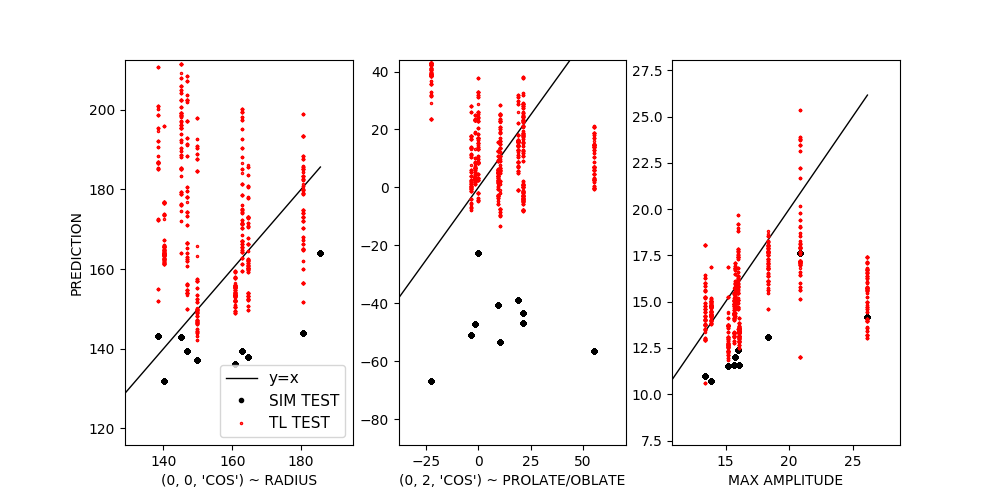}
  \caption{Similar to Figure \ref{fig:xval_data_images_lg} but for the simple linear output calibration instead of transfer learning.}
  \label{fig:xval_data_images_lg_simple_TL}
  \end{center}
\end{figure}

Once we have demonstrated that transfer learning systematically improves predictions of the simulation-trained model, it is natural to ask if a similar improvement could be achieved using a simpler calibration method. If the objective is to improve the predictions, do we really need to design a new surrogate model? Perhaps the correction in the outputs could be determined independently of the inputs.

Such an approach was successfully applied by \cite{varchas} to ICF experiments at the Omega laser facility at the University of Rochester. Our data, measured at NIF, however, are in many ways more challenging than the Omega data. Lasers at NIF have a lower repetition rate hence we have fewer shots available for the calibration. Predictions at NIF are also more difficult in that the high-resolution capsule simulation captures only a part of the experiment, and the inference of the unsimulated first part of the experiment adds uncertainty to the simulation inputs. There is no such uncertainty at Omega, where the lasers shine directly at the target capsule target. Finally, we attempt to calibrate multi-modal data, while \cite{varchas} calibrated only the scalars. Our results cannot therefore be easily compared against those described in \cite{varchas}. Subsequently to our work, an attempt to circumvent some of the aforementioned challenges has been made by calibrating scalar outputs from low-fidelity simulations that capture the entire NIF experiment \cite{kelli_hohl}. Expanding that workflow to multi-modal data could be a natural extension of that research but would require a significant effort.

In this paper, to formulate a baseline for transfer learning, we design a simple output calibration method:

\begin{equation} \label{eq:linear}
y' _{EXP} = L (y'_{SIM}).
\end{equation}

In this equation $L$ stands for a linear mapping between simulated and experimental outputs. Since we have fewer than 10 experiments available to determine $L$, we do not even consider a non-linear mapping. With so few training samples, we also cannot determine this mapping for raw outputs consisting of 10 scalars and 60x60 pixels; this would require determination of an overwhelming number of coefficients. Instead, equation \ref{eq:linear} incorporates $y'$: a compressed version of the outputs $y$ computed as:

\begin{equation} \label{eq:compress}
y\prime = T(y),
\end{equation}

with

\begin{equation} \label{eq:compress2}
T = PCA_{all}  [PCA_{img}(y_{image}), y_{scalars}],
\end{equation}

where $PCA$ stands for the matrix of coefficients determined by the principal component analysis, or, in other words, the first $k$ eigenvectors of the covariance matrix $yy^T$ with $k$ indicating the size of the representation after compression. Equation \ref{eq:compress2} means that images are first compressed using $PCA_{img}$ to a small number of principal components. This prevents assigning unreasonably high weight to the images relative to scalars in the subsequent regression. Magnitudes of these components are also normalized to be in the same range as scalars. The final compression $PCA_{all}$ gives an option to further reduce the number of components as an attempt to ensure that the linear regression $L$ is not ill-conditioned. The operator $T$ is estimated from over 90,000 simulated outputs, and then applied to compress the experimental data as well. Coefficients of the linear mapping $L$ are determined by comparing the compressed seven training experiments $y'_{EXP}$ with the compressed predictions $y'_{SIM}$ of the simulation-trained surrogate.

We carried out extensive hyperparameter testing to find the optimal number of the $PCA_{all}$ components, $PCA_{img}$ components, the strength of the standard L2 regularization, and at which steps the normalization is beneficial. It turns out that using $PCA_{all}$ with fewer then 10 components does not allow us to calibrate even the training samples. In such case, the linear model simply does not have enough capacity to represent all outputs because there is not enough linear correlation between different outputs to allow efficient compression.

We obtained the best simple calibration results by compressing the images using four principal components and then concatenating them with the ten scalars without further compression $PCA_{all}$. The simple calibration model is, however, expected to generalize less well than the transfer learned model for two reasons. First, the transformation $L$ is linear, while transfer learning can capture nonlinear transformations. Second, unlike transfer learning, the simple calibration ignores the relationship of the outputs with the inputs, and relies on an arbitrary regularization.



This expectation is confirmed in Figure \ref{fig:bagging_all}. Here, the setup is slightly different than in Section \ref{sec:results} in that each of the 10 shots is predicted only 15 times (instead of 500) by randomly selected seven shots. Not only are the predictions of the simple calibration much worse than the transfer learned predictions but the simple calibration predictions are worse than the predictions of the initial model for eight out of 10 scalars, sometimes by a large margin. Transfer learning, on the other hand, improved the initial predictions for the majority of the scalar outputs.

To investigate whether simple calibration could be improved by bootstrap aggregation, we averaged out the 15 predictions for each holdout shots and plotted the means in Figure \ref{fig:bagging_means}. Many poorly predicted outliers were eliminated by this process but simple calibration still degraded the fit of the simulation-trained model for seven out of 10 scalars.

As a side note, in Figures \ref{fig:bagging_all} and \ref{fig:bagging_means} we switched from using a $\chi^2$ metric in the transfer learning optimization to the simple $L2$ minimization but the $\chi^2$ metric is still used to evaluate the predictions. Regardless of this change, transfer learning still improved the fit for the majority of the scalars. It is interesting, however, to explain the degradation in the fit for one of the scalars: the DT\_YIELD. The prediction error of the simulation-trained model for DT\_YIELD is so small compared to other scalars, that the $L2$ optimization, targeting mainly the large errors, allowed for some degradation in the fit of DT\_YIELD. In Section \ref{sec:results}, however, when $\chi^2$ was used to measure the goodness of fit, the relatively small measurement error of DT\_YIELD gave this scalar a higher weight in the optimization leading to the improved predictions.

For X-ray images, to enable a direct comparison of simple calibration with the transfer learning results in Figure \ref{fig:xval_data_images_lg}, we computed 500 simple calibration models by randomly splitting the 10 experiments into 7 training and 3 validations shots. The results are shown in Figure \ref{fig:xval_data_images_lg_simple_TL}. Simple calibration moved the predictions of the three image characteristics towards the experimental values but the spread of the predictions in much larger than in transfer learning predictions. In particular, many of the radius predictions are worse than the predictions of the initial model.

In summary, the simple output calibration degraded predictions of the simulation-trained model for the majority of the scalars, and introduced a large uncertainty in the predictions of the images.

\section{Conclusions and future work}
\label{sec:conclusions}

We have designed a transfer learning method for a complicated regression problem: a calibration of the simulation-trained multi-modal surrogate model against very few experiments. For the first time, we included multiple output data types in the calibration and demonstrated a successful application using only seven ICF experiments to train the model. Transfer learning systematically improved predictions of the simulation-trained surrogate for both real and synthetic validation data; the latter were carefully designed to mimic the simulation bias. Such improvement was not possible with a simple calibration of the outputs (a method we have designed as a baseline), which ignores the experimental inputs. This suggests that the relationship between inputs and outputs that is learned from the simulations, and leveraged by transfer learning, may help to regularize the underdetermined calibration problem.

Extending the transfer learning code to work with an arbitrary set of modalities may not be trivial. New types of data will likely require new, specialized architectures of the autoencoder. The idea of retraining the decoder may, however, can be reused in other architectures.

Future work may involve the integration of transfer learning with the Bayesian estimation of model parameters~\cite{jimpop} to match the experimental outputs as consistently with the simulation code as possible. Recent advances in calibrating uncertainty estimates~\cite{anderson} could also be incorporated in transfer learned models. Finally, experimenting with the cutting-edge methods of transfer learning \cite{mimicgan} could potentially further  improve the generalization of the calibrated models.


\section*{Acknowledgments}
We thank Art Pak and Benjamin Bachmann for help with experimental images, and Kelli Humbird and Luc Peterson for discussions about transfer learning.
The authors appreciate comments from an anonymous reviewers, which were helpful in improving the manuscript.

This work was performed under the auspices of the U.S. Department of Energy by Lawrence Livermore National Laboratory under Contract DE-AC52-07NA27344 and was supported by the LLNL-LDRD Program under Project No. 18-SI-002. Released as LLNL-JRNL-829622.
This document was prepared as an account of work sponsored by an agency of the United States government. Neither the United States government nor Lawrence Livermore National Security, LLC, nor any of their employees makes any warranty, expressed or implied, or assumes any legal liability or responsibility for the accuracy, completeness, or usefulness of any information, apparatus, product, or process disclosed, or represents that its use would not infringe privately owned rights. Reference herein to any specific commercial product, process, or service by trade name, trademark, manufacturer, or otherwise does not necessarily constitute or imply its endorsement, recommendation, or favoring by the United States government or Lawrence Livermore National Security, LLC. The views and opinions of authors expressed herein do not necessarily state or reflect those of the United States government or Lawrence Livermore National Security, LLC, and shall not be used for advertising or product endorsement purposes.


\vspace{10pt}

\bibliography{paper.bbl}

\end{document}